\documentclass{svproc}
\usepackage{url}
\usepackage{graphicx} 
\providecommand{\subparagraph}{}
\usepackage{pythonhighlight}
\usepackage{hyperref}

\begin{document}
\mainmatter              
\title{UAIbot: Beginner-friendly web-based simulator for interactive robotics learning and research}
\titlerunning{UAIbot}  
\author{Johnata Brayan\inst{1} \and Armando Alves Neto\inst{1} \and Pavel Petrovi\v{c}\inst{2} \and Gustavo M Freitas\inst{1} \and Vinicius Mariano Gon\c{c}alves\inst{1}}
\authorrunning{Brayan et al.} 
\institute{Graduate Program in Electrical Engineering - Universidade Federal de Minas Gerais - Av. Antônio Carlos 6627, 31270-901, Belo Horizonte, MG, Brazil
\and
Comenius University, Bratislava, Slovakia}

\maketitle              

\begin{abstract}
This paper\footnote{We acknowledge the support of the funding agencies CAPES, CNPQ, FAPEMIG (under grant APQ-02144-18).} presents \emph{UAIbot}, a free and open-source web-based robotics simulator designed to address the educational and research challenges conventional simulation platforms generally face. The Python and JavaScript interfaces of UAIbot enable accessible hands-on learning experiences without cumbersome installations. By allowing users to explore fundamental mathematical and physical principles interactively, ranging from manipulator kinematics to pedestrian flow dynamics, UAIbot provides an effective tool for deepening student understanding, facilitating rapid experimentation, and enhancing research dissemination.

\end{abstract}

\begin{figure}[h]
    \centering
    \includegraphics[width=1.0\textwidth]{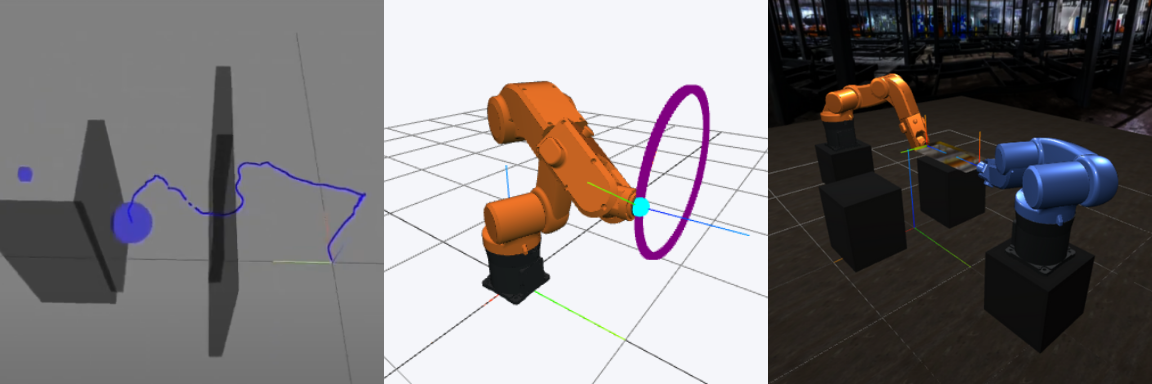}
    \caption{ Simulations made with UAIbot.}
    \label{fig:uaibot_hero_image}
\end{figure}

\section{Introduction}

Robotic simulation is essential in robotics education, research, and rapid prototyping. Despite the proven utility of established desktop-based simulators such as CoppeliaSim \cite{coppeliaSim}, Gazebo \cite{gazebo}, and MATLAB toolboxes \cite{matlabtoolbox}, the challenges related to their complexity of installation and shallow learning curve have hindered widespread adoption in certain niches. Furthermore, it's important to note that MATLAB is a paid software, which can be a barrier for some users. Currently, the rise of web-based tools and cloud platforms highlights the need for simulators that are accessible and seamlessly integrated into contemporary workflows. 
We introduce \emph{UAIbot} (Fig.~\ref{fig:uaibot_hero_image}), a free and open-source multilingual web-based robotics simulator designed to meet these requirements. The Python and JavaScript UAIbot interfaces facilitate a streamlined simulation setup, allowing the testing of control strategies and enabling the sharing of interactive scenarios. Accessible through Google Colab or local environments, UAIbot supports various robot types and offers a lightweight low-level interface for beginners and experts. The user-friendly nature of the system is intended to promote hands-on experimentation. Additionally, by providing transparent access to the underlying mechanics, UAIbot enhances comprehension and accelerates iterative development within robotics research and education. 

This paper first reviews the pertinent literature, presents an overview of existing robotics simulators and details their strengths and limitations. Subsequently, the Architecture section delineates the design of the UAIbot system, underscoring browser-based integration and a versatile computational environment. The Features section follows by explaining the intuitive API and interactive 3D interface of UAIbot. It highlights multilingual support and demonstrates how UAIbot addresses accessibility and ease of use. Next, the Use Cases section illustrates the practical applications of UAIbot, ranging from classroom instruction to research data visualization. We describe a cooperation with Comenius University including adding support for interaction of UAIbot with a real robot. The conclusion explores potential avenues for future development and enhancements to this open-source web-based robotics simulation platform.

\section{Related Work}

Robot simulation is a valuable task across various domains, leading to the development of numerous simulation tools. Despite this proliferation, there is a compelling need for a new solution, particularly to examine the mathematical and physical principles in the modeling and control of robotic manipulators. This paper delineates the specific requirements and context that led to the development of our proposed simulation platform.

In general, existing solutions can be categorized into three broad groups. Firstly, industrial solutions provided by companies such as ABB, Fanuc, KUKA, and Universal Robots include proprietary tools and programming languages, e.g., ABB RAPID, Fanuc KAREL, and KUKA KRL. These are usually integrated with commercial simulators like RoboDK, ABB RobotStudio, and Fanuc Roboguide, allowing rapid application-oriented development and simulation. However, these tools are designed for high-level abstractions, are closed-source, paid, and lack flexibility in exploring the fundamental mathematical and physical concepts essential for pedagogical objectives in the field of robotic manipulators.

In contrast, academic solutions emphasize deeper engagement with the core principles. Open-source frameworks such as ROS, used in conjunction with simulators like Gazebo, Webots, and CoppeliaSim, promote detailed exploration and the ability to implement custom algorithms. These environments are free and support the investigation into the mathematical and physical foundations of robotics but present challenges due to their complexity. The process of installing packages, managing dependencies, and navigating extensive functionalities can hinder the focus on the core theoretical concepts. This complexity poses an entry barrier for introductory courses dedicated to modeling and control basics.

Third, educational simulators often use highly abstract programming languages and simplified interfaces. Notable examples include Scratch, EV3, and ROBOTC, coupled with simulation environments such as Robosim, OpenRoberta, or Robot Virtual Worlds. Although effective for beginners, these tools obfuscate the mathematical and physical complexities critical for advanced learners seeking to understand fundamental robotics principles. Therefore, while they facilitate early-stage learning, they do not meet the needs of those who are looking to grasp the basic concepts of robotics.

Our analysis identifies an untapped niche: a freely accessible open-source simulation platform that offers ease of installation and provides low-level control over simulations. Such a tool would enable students and researchers to explore theoretical concepts without the constraints of high-level abstractions or intricate software installations present in existing solutions. Addressing this gap drives the development of our platform.

\section{Architecture}

UAIbot has two versions: UAIbotPy\footnote{\href{https://uaibot.github.io/docs/User_Guide/UAIbotPy_quickstart.html}{https://uaibot.github.io/docs/User\_Guide/UAIbotPy\_quickstart.html}} and UAIbotJS\footnote{\href{https://uaibot.github.io/docs/User_Guide/UAIbotJS_quickstart.html}{https://uaibot.github.io/docs/User\_Guide/UAIbotJS\_quickstart.html}}. The UAIbotPy architecture is designed to integrate a Python-based simulation backend with a flexible 3D visualization front end that is based on the browser. This approach takes advantage of modern web technologies, providing users with an interactive and portable environment that can be accessed online or run locally. The underlying philosophy is to maintain a clear separation between high-level simulation logic in Python and the low-level rendering details handled by JavaScript and WebGL through the Three.js library.

The core of the system is the Simulation class, which orchestrates the simulation environment. It manages a collection of virtual objects, ranging from simple primitives such as boxes or point lights to more complex entities, and provides user-defined parameters that shape the visual and interactive behavior of the simulation. The key attributes controlled by the class include ambient light intensity, background coloring, viewport dimensions, camera configuration, and scene composition. 

Internally, the Python layer handles scene construction and configuration, generating a structured representation of the environment and its objects. To visualize the result, the Simulation class exports the scene definition into a self-contained HTML/JavaScript code snippet. This code leverages Three.js to handle 3D rendering and interactivity directly in the user\textsc{\char13}s browser. By calling the simulation run method (to visualize the simulation inside a Jupyter notebook) or the save method (to save it locally as an HTML file), users can view and interact with their scenario in real-time, adjusting the simulation timeline, manipulating objects, and observing robotic behavior from multiple perspectives.

UAIbotJS is a simplified version of UAIbotPy written entirely in JavaScript. It adopts a client-side approach using JavaScript, Three.js, and Math.js to run robotics simulations directly in the browser. At its core is the Simulation class, which sets up a WebGL-rendered scene with a camera, lighting, axes, and a grid. Users can easily manipulate their viewpoint via OrbitControls, making the environment more intuitive. The simulation of open-chain robotics manipulators is managed by the Robot class, which allows users to define manipulators via Denavit-Hartenberg parameters, configure joint angles, and calculate both forward and inverse kinematics. UAIbotJS is written as an ES6 module and can be easily imported into any JavaScript project via a CDN.

\section{Features}

A notable feature shared by both UAIbotPy and UAIbotJS lies in the extensive documentation\footnote{\href{https://uaibot.github.io/}{https://uaibot.github.io/}} accompanying these tools. Their creators have taken care to present not only code-level references and API specifications but also conceptual explanations and illustrative examples. By providing clear guidance, relevant tutorials, and well-structured resources, the documentation empowers users, whether beginners or experienced robotics professionals, to navigate the platform’s features effectively. Ultimately, this has the potential to facilitate the learning curve, accelerate troubleshooting, and encourage experimentation.

\subsection{UAIbotPy}

UAIbotPy provides a comprehensive Python-based interface for creating, visualizing, and manipulating robotic simulations. Its architecture and API aim to balance ease of use with the flexibility required by advanced users. Key features include:

\subsubsection{Object Library:}
UAIbotPy offers a collection of simulation-ready objects. Users can instantiate and manipulate primitive geometric shapes, such as spheres, boxes, cones, and cylinders, along with more complex elements such as 3D models and point clouds. These objects can be combined into groups, integrated into robot assemblies, or used as collision obstacles, enabling users to prototype complex scenarios.

The library supports a variety of robot models, including, but not limited to, ABB CRB, DARWIN MINI, KUKA KR5, EPSON T6, FRANKA ERMIKA 3, STAUBLI TX60, and KUKA LBR IIWA.
Each robot model is implemented as an object that can be integrated into simulations, providing users with a flexible and extensible toolkit for robotics research and development.

\subsubsection{Robotic Modeling Tools:}
Central to UAIbotPy are classes and methods dedicated to robotic systems. Developers can specify manipulators using the Denavit-Hartenberg convention, construct kinematic chains, and integrate multiple links, joints, and 3D models to form robotic arms. The built-in forward and inverse kinematics, Jacobian computation, and dynamic modeling tools allow users to perform analysis and control design in different robotic manipulators. UAIbot has no physics engine, but dynamical simulations can be performed with the help of functions that implement the canonical equation \cite{Featherstone2008}. Furthermore, functionality such as constraint-based motion planning, collision avoidance, and joint-limit enforcement provide a foundation for more advanced control techniques.

\subsubsection{Pedestrian Simulation Tools:}
\begin{figure}[h]
    \centering
    \includegraphics[width=0.7\textwidth]{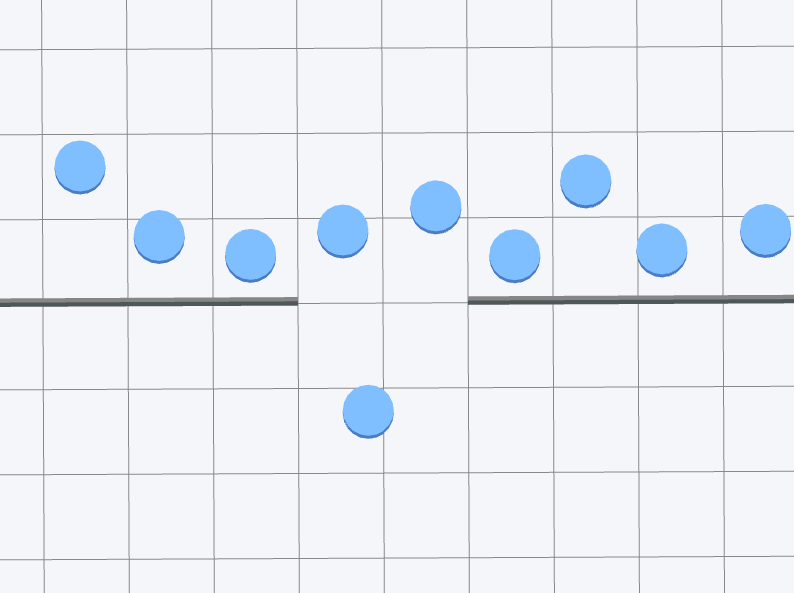}
    \caption{ Simulation of Pedestrians Evacuating a room made with UAIbot.}
    \label{fig:uaibot_evacuation}
\end{figure}

In addition to robotic manipulators, UAIbotPy supports the simulation of pedestrian behavior through the social force model, enabling human-centered robot navigation research. Users can adjust parameters such as speed, interaction radii, and goal waypoints to generate realistic pedestrian trajectories, as shown in Fig.~\ref{fig:uaibot_evacuation}. In this simulation\footnote{\href{https://uaibot.github.io/assets/simf.html}{https://uaibot.github.io/assets/simf.html}} the platform has been used to simulate room evacuations, showcasing its ability to model complex human dynamics.

\subsubsection{Material Control:}
UAIbotPy leverages Three.js-based rendering to produce customizable visuals. The MeshMaterial class grants fine-grained control over surface properties, including reflectivity, roughness, transparency, and more. Textures and environmental maps can be applied to create realistic settings. By experimenting with lighting, background colors, and environment maps (LDR/HDR), users can tailor the appearance of the scene, thus enhancing the interpretability of simulation results. Users can also import custom 3D models.

\subsubsection{Interactive Simulations and Scenes:}
The Simulation class simplifies the creation and management of complex scenes. Users can configure camera parameters, lighting intensity, and grid visibility, as well as add arbitrary objects and robots directly to the environment. Various methods enable "factories" in the environment that generate common settings with minimal effort. Simulations can be dynamically updated, paused, resumed, and traversed over time, allowing users to inspect robot motions, verify control strategies, or highlight specific moments in a demonstration.

\subsubsection{Seamless Integration and Sharing:}
Simulations can be rendered inline in \textit{Jupyter Notebooks} or \textit{Google Colab}, facilitating rapid prototyping and sharing through interactive web-based notebooks. The entire simulation can be exported as a self-contained HTML file, enabling educators and researchers to distribute and present their scenarios without requiring users to install or configure additional software.

\subsubsection{Utility Functions for Transformation and Computation:}
UAIbotPy includes a set of utility functions, accessible through the \texttt{uaibot.Utils} class, to simplify common robotics-related computations. These functions help users avoid repetitive coding of fundamental operations and facilitate a clearer focus on higher-level robotics tasks.
\begin{itemize}
\item \textbf{Matrix and Vector Operations:} Functions like \texttt{dp\_inv(mat, eps)} compute the damped pseudoinverse of a matrix, while \texttt{jac(f, x, delta)} numerically approximates the Jacobian of a vector-valued function. The \texttt{S(v)} function generates a skew-symmetric matrix from a 3D vector, allowing computation of cross-products through matrix multiplication.

\item \textbf{Homogeneous Transformations and Rotations:} The utilities \texttt{inv\_htm(htm)} efficiently compute the inverse of a homogeneous transformation matrix, while functions such as \texttt{rot(axis, angle)}, \texttt{rotx(angle)}, \texttt{roty(angle)}, and \texttt{rotz(angle)} generate rotation matrices about specified axes. The \texttt{trn(vector)} function produces translation transformations. These building blocks simplify the construction and manipulation of 3D transformations, which is critical for analyzing robot kinematics, grasp poses, and sensor alignments.

\item \textbf{Geometric and Kinematic Tools}: With \texttt{euler\_angles(htm)}, users can extract roll, pitch, and yaw angles from a given transformation, helping in tasks like orientation visualization and sensor alignment. Random transformations can be generated using (\texttt{htm\_rand(trn, rot)}) for simulation scenarios, Monte Carlo analyses, or testing against variations in robot configuration.
\end{itemize}

\subsection{UAIbotJS}
UAIbotJS provides an intuitive browser-based platform for visualizing and experimenting with robotic systems directly within a Web environment. Built atop JavaScript and Three.js, it delivers interactivity without requiring complex installations or specialized local software. Key features include:

\subsubsection{Scene Setup and Management:}
The \texttt{Simulation} class forms the backbone of the creation of the environment by UAIbotJS. By instantiating a \texttt{Simulation} object, users can easily set up a 3D scene, adjust the camera perspective, control lighting, and render animations. Methods like \texttt{setAnimationLoop} simplify real-time simulation updates, while \texttt{fitWindow} ensures the visualization dynamically adapts to browser window resizing.

\subsubsection{Object Library:}
Core geometric primitives \texttt{Ball}, \texttt{Box}, \texttt{Cylinder}, and \texttt{Frame} inherit from a common \texttt{Objsim} base class. This design ensures a consistent interface for object manipulation. Each object supports the \texttt{setHTM} method, allowing straightforward positioning and orientation using a homogeneous transformation matrix. Users can quickly create and transform these objects, enabling quick prototyping of robotic work cells or educational demonstrations.

\subsubsection{Robotic Manipulator Modeling and Control:}
The \texttt{Robot} class enables the modeling of generic open-chain robotic manipulators using Denavit-Hartenberg (DH) parameters. Once defined, users can compute forward kinematics (\texttt{fkm}), retrieve the geometric Jacobian (\texttt{\_jac\_geo}), and adjust joint configurations with \texttt{config}. The ability to 'catch' and 'release' scene objects adds interactive elements to test pick-and-place tasks. Predefined helper methods like \texttt{create\_kuka\_kr5} and \texttt{create\_epson\_t6} offer instant access to well-known robot models, facilitating comparisons and benchmarking.

\subsubsection{Direct Interaction and Real-Time Updates:}
Since UAIbotJS runs directly in the Web browser, no additional software installation is required. Users can integrate simulations into webpages, making demonstrations and learning modules readily accessible. With interactively displayed objects and robots, students and researchers can pause, inspect, and adjust scenarios in real-time, gaining insight into robotic configurations, kinematics, and motion strategies.

\subsubsection{Utility Functions for Transformation and Computation:}
A suite of helper functions (\texttt{rot}, \texttt{rotx}, \texttt{roty}, \texttt{rotz}, \texttt{trn}, \texttt{s}, \texttt{dp\_inv}) simplifies common robotic tasks. Users can easily generate and apply rotations, translations, cross-product matrices, and damped pseudoinverses. These tools streamline the development of custom algorithms for control, path planning, and motion analysis.

\section{Use Cases}

The design of UAIbot supports a wide range of applications in both educational and research settings. Whether integrated into web-based presentations or employed to clarify complex theoretical ideas, UAIbot offers the possibility of engaging, hands-on experiences that can benefit learners and researchers alike.

\begin{figure}[h]
    \centering
    \includegraphics[width=1\textwidth]{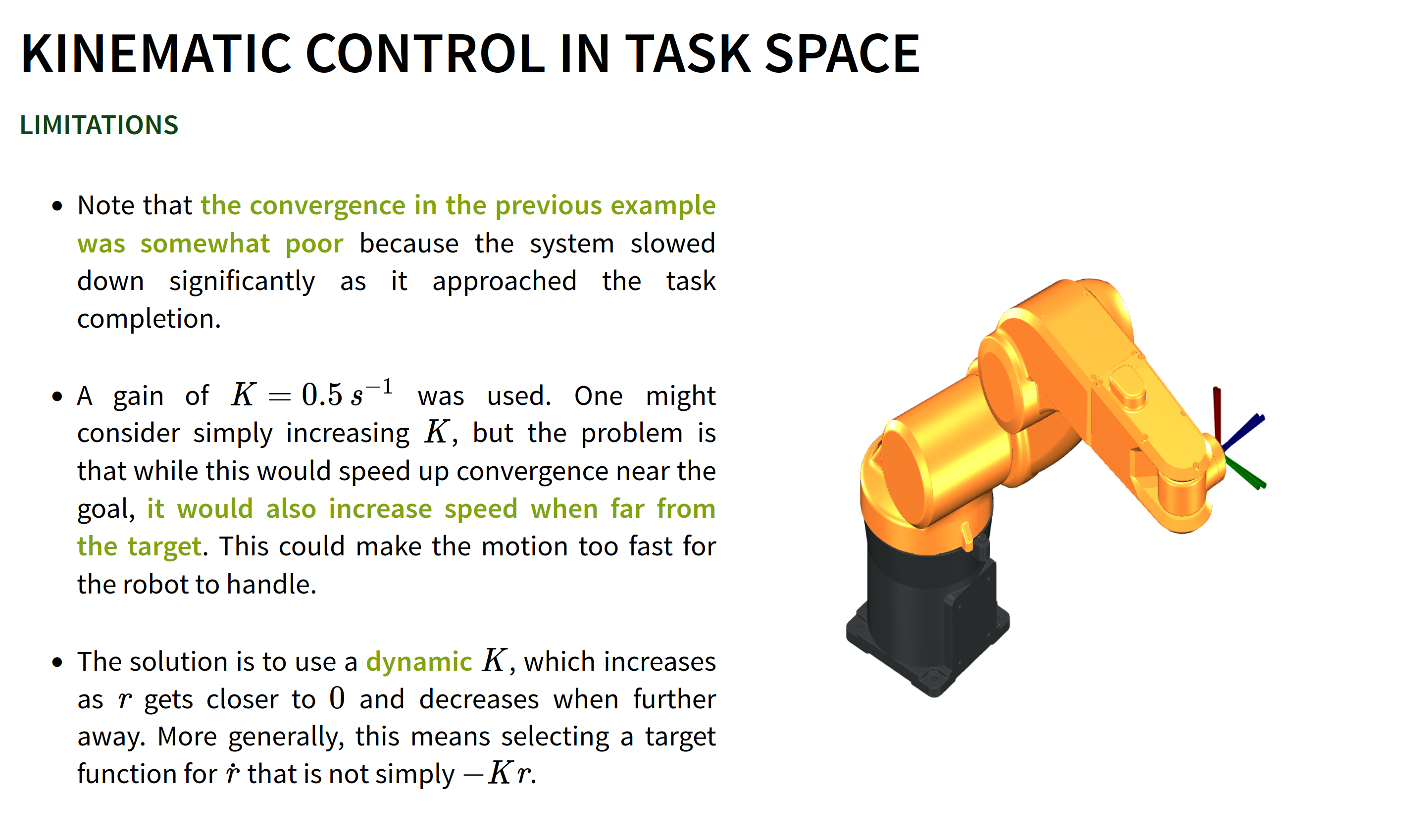}
    \caption{ Slides presentation using a UAIbot simulation during a lesson.}
    \label{fig:uaibot_pres}
\end{figure}

\subsection{Interactive Lessons and Conceptual Demonstrations}
UAIbot simulations can be saved as standalone HTML files, making them embeddable into online presentations\footnote{\href{https://uaibot.github.io/class_presentations/}{https://uaibot.github.io/class\_presentations/}} or lecture notes. For example, instructors can integrate a robotics simulation into a slideshow created with any HTML slideshow processor (e.g. slides.com), allowing students to manipulate and view the scenario in real-time. This direct interaction helps maintain participation and enhances the understanding of robotics principles during a lesson, as illustrated in Fig.~\ref{fig:uaibot_pres}. 

UAIbot has been used for more than three years in Professor Vinicius Mariano's robotic manipulator course at the Federal University of Minas Gerais. The platform allows him to embed animations directly into his course material slides, which are presented in HTML format. This integration not only brings theoretical concepts to life, but also provides students with hands-on interaction that deepens their learning experience.

\begin{figure}[h]
    \centering
    \includegraphics[width=0.7\textwidth]{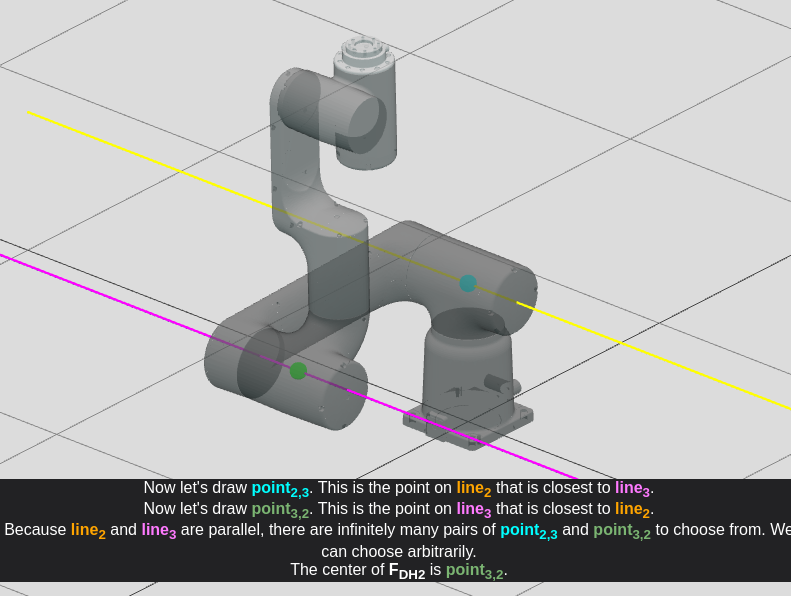}
    \caption{ Example of D-H referential demonstration.}
    \label{fig:uaibot_d_h}
\end{figure}

Beyond interactive lessons, UAIbot is particularly effective for conceptual demonstrations. Many concepts, especially those involving spatial reasoning such as the Denavit-Hartenberg (DH) convention, are challenging to convey through static diagrams and formulas. The DH convention is known for its ambiguous terminology, specialized indexing, and complex application \cite{teach-D-H}. UAIbot addresses these challenges by providing students with an animated 3D simulation\footnote{\href{https://uaibot.github.io/docs/Theoretical\_Reference/denavit\_hartenberg\_convention.html}{https://uaibot.github.io/docs/Theoretical\_Reference/denavit\_hartenberg\_convention.html}}, as seen in Fig.~\ref{fig:uaibot_d_h}, where they can rotate the robot and navigate through time while the convention is applied. By interactively visualizing D-H frames and transformations, students gain deeper insights, reduce confusion, and accelerate the learning process.

\subsection{Data Visualization in Research}

\begin{figure}[h]
    \centering
    \includegraphics[width=1.0\textwidth]{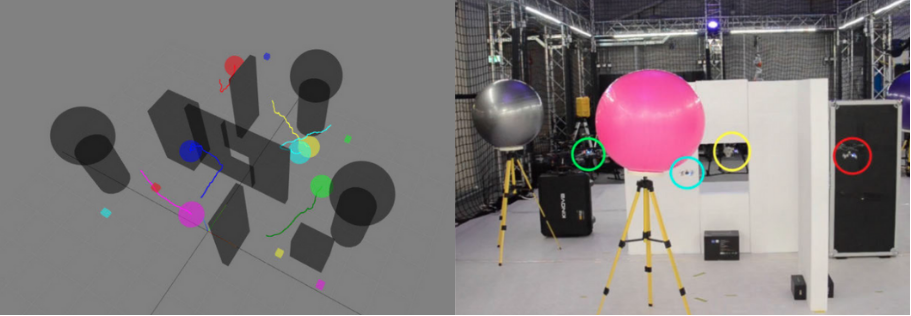}
    \caption{ Left: UAIbot simulation, Right: Experimental setup. Images taken from \cite{dataviz}.}
    \label{fig:uaibot_dataviz}
\end{figure}

The ability of UAIbot to display and animate time-varying three-dimensional data also makes it a practical tool for data visualization in research and experimentation, as seen in \cite{10.1109/TRO.2024.3400924}, \cite{GONCALVES2024104601}, and \cite{dataviz}. Fig.~\ref{fig:uaibot_dataviz} shows an example \cite{dataviz} of the proposed simulator used for this exact purpose. An experiment was set up to test a novel navigation technique for drones, and the position of the drones in time was recorded with the help of reflexive markers. These recorded positions were then used to animate the path covered by the drones inside a UAIbot simulation. The drones and the environment were represented as geometric primitives. This made it easy to visually compare the performance of the proposed technique with the old baseline. 

\subsection{Integration of UAIbot with a Real Robot}

\begin{figure}[h]
    \centering
    \includegraphics[width=1.0\textwidth]{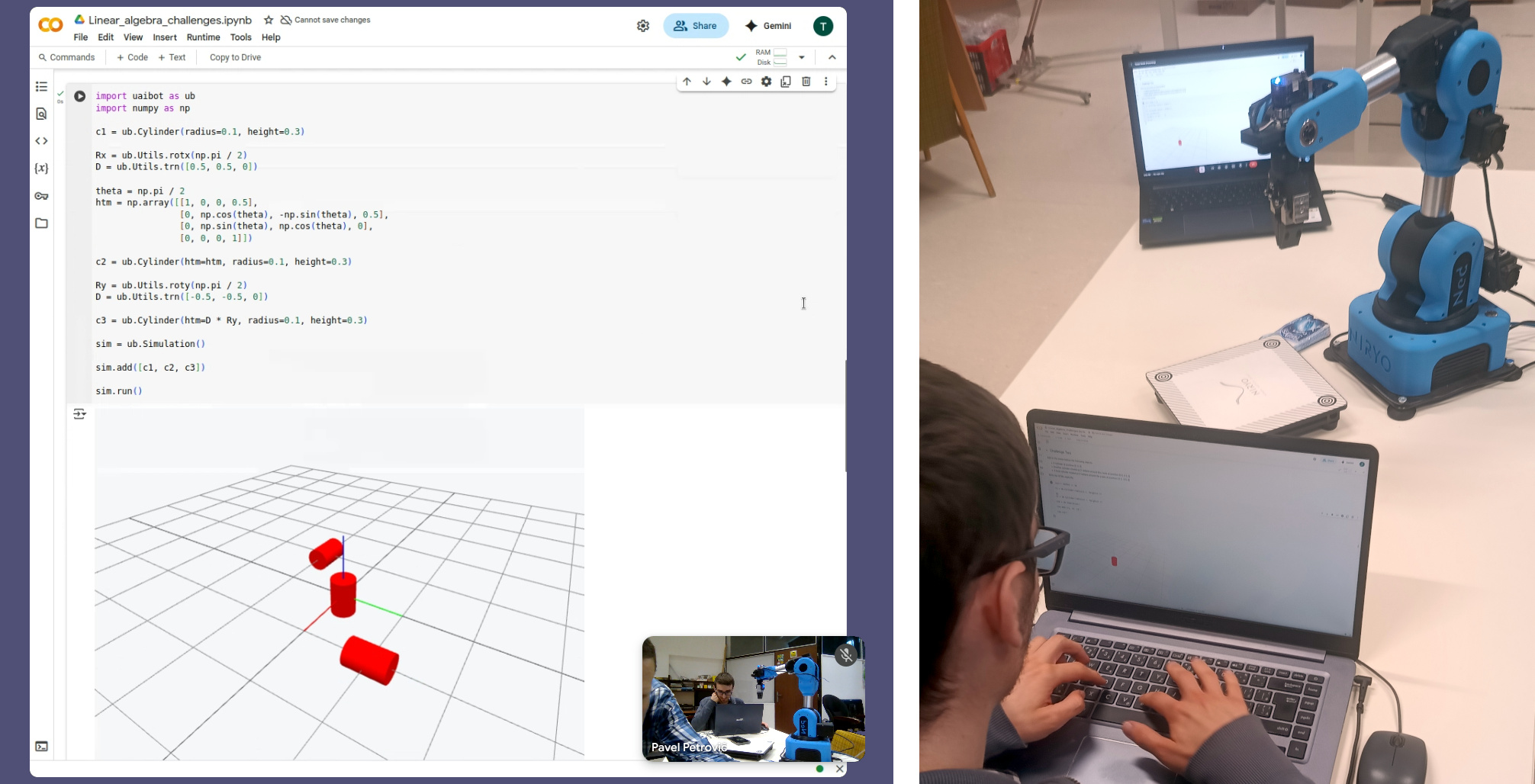}
    \caption{ Using UAIbot and Niryo Ned in a robotics course at Comenius University in Slovakia with on-line support from Brazil.}
    \label{fig:niryo}
\end{figure}

Students at the Faculty of Mathematics, Physics and Informatics, Comenius University in Bratislava had the opportunity to be introduced into UAIbot simulator by its author on several occasions (three different courses in two different years, about 5 sessions in total, both virtual and live versions, including both theoretical explanations and practical hands-on exercises), see Fig.\ref{fig:niryo}. This ignited the idea of integrating JavaScript version of UAIbot with a real robot Niryo Ned that we normally use in some of the robotics related courses. On one hand, we have included Niryo Ned in the set of robots supported by JavaScript version of UAIbot. On the other hand, we have developed a software prototype that allows simultaneous operation of both the simulated and real robot. It is achieved by serving the JavaScript version of UAIbot from a Python webserver and websocket connection. The Python program controls the Niryo Ned robot using the official pyniryo library, but it allows sending a request to the simulator to move the simulated robot to a desired pose - either formulated in the joints configuration space or in the world coordinates space. Similarly, our prototype allows a JavaScript program that utilizes the UAIbot functionality directly to also connect to the Python program to control the real robot as desired\footnote{\href{https://github.com/Robotics-DAI-FMFI-UK/niryoned}{https://github.com/Robotics-DAI-FMFI-UK/niryoned}}.

\section{Conclusion}

UAIbot stands as a versatile and accessible tool for robotics education and research. Bridging the gap between highly abstract or proprietary simulators and the need for low-level fundamental understanding, it empowers educators, students, and researchers to engage more deeply with robotic concepts. The ease of integration of UAIbot, the interactive simulations, and the open nature encourage exploration of established robotics principles and emerging research areas.

By delivering accessible, interactive, and visually rich simulations, UAIbot empowers educators and researchers to create engaging lessons, clarify intricate robotics concepts, and analyze complex datasets more intuitively.

As the platform evolves, future work will focus on expanding robot model libraries, enhancing support for advanced control algorithms, and integrating additional modules for specialized domains, such as multirobot coordination or augmented reality interfaces. In doing so, UAIbot aims to continue shaping how learners and experts approach robotics modeling, control, and analysis, ultimately enriching the educational landscape and accelerating innovations in robotics research.

\section*{Acknowledgement}
This study was financed in part by the Coordena\c{c}\~{a}ço de Aperfeiç\c{c}oamento de Pessoal de N\'{i}ível Superior - Brazil (CAPES) - Finance Code 001

\bibliographystyle{plain}
\bibliography{references}

\begin{thebibliography}{1}

\bibitem{matlabtoolbox}
Peter Corke.
\newblock {\em Robotics, Vision and Control: Fundamental Algorithms in MATLAB}.
\newblock Springer Publishing Company, Incorporated, 1st edition, 2013.

\bibitem{Featherstone2008}
Roy Featherstone and David~E. Orin.
\newblock {\em Dynamics}, pages 35--65.
\newblock Springer Berlin Heidelberg, Berlin, Heidelberg, 2008.

\bibitem{10.1109/TRO.2024.3400924}
Vinicius~Mariano Gon\c{c}alves, Anthony Tzes, Farshad Khorrami, and Philippe Fraisse.
\newblock Smooth distances for second-order kinematic robot control.
\newblock {\em Trans. Rob.}, 40:2950–2966, May 2024.

\bibitem{GONCALVES2024104601}
Vinicius~Mariano Gonçalves, Dimitris Chaikalis, Anthony Tzes, and Farshad Khorrami.
\newblock Safe multi-agent drone control using control barrier functions and acceleration fields.
\newblock {\em Robotics and Autonomous Systems}, 172:104601, 2024.

\bibitem{dataviz}
Vinicius~Mariano Gonçalves, Prashanth Krishnamurthy, Anthony Tzes, and Farshad Khorrami.
\newblock Control barrier functions with circulation inequalities.
\newblock {\em IEEE Transactions on Control Systems Technology}, 32(4):1426--1441, 2024.

\bibitem{gazebo}
N.~Koenig and A.~Howard.
\newblock Design and use paradigms for gazebo, an open-source multi-robot simulator.
\newblock In {\em 2004 IEEE/RSJ International Conference on Intelligent Robots and Systems (IROS) (IEEE Cat. No.04CH37566)}, volume~3, pages 2149--2154 vol.3, 2004.

\bibitem{coppeliaSim}
E.~Rohmer, S.~P.~N. Singh, and M.~Freese.
\newblock Coppeliasim (formerly v-rep): a versatile and scalable robot simulation framework.
\newblock In {\em Proc. of The International Conference on Intelligent Robots and Systems (IROS)}, 2013.
\newblock www.coppeliarobotics.com.

\bibitem{teach-D-H}
Dariusz Zarychta and Igor Zubrycki.
\newblock Methodology for teaching the denavit-hartenberg notation.
\newblock In Richard Balogh, David Obdr{\v{z}}{\'a}lek, and Martin Fislake, editors, {\em Robotics in Education}, pages 126--137, Cham, 2024. Springer Nature Switzerland.

\end{thebibliography}

\end{document}